\documentclass{article}
\usepackage{hyperref}
\usepackage{url}
\usepackage{booktabs}
\usepackage{amsfonts}
\usepackage{amssymb}
\usepackage{amsmath}
\usepackage{graphicx}
\usepackage{float}
\usepackage{natbib}
\usepackage{adjustbox}
\usepackage[utf8]{inputenc}
\usepackage[T1]{fontenc}
\usepackage{lmodern}
\usepackage[margin=1in]{geometry}
\usepackage{indentfirst}
\usepackage{caption}

\title{Exogenous Dropout: A Simple, Strong Baseline for Corruption-Robust Time Series Forecasting with Covariates}

\author{%
\begin{minipage}{0.96\textwidth}
\centering
HU Hao\textsuperscript{1} \qquad AI Xue-shan\textsuperscript{1,2,\,*}\\[0.8ex]
{\normalsize\textsuperscript{1}State Key Laboratory of Water Resources Engineering and Management, Wuhan University, Wuhan 430072, China}\\[0.2ex]
{\normalsize\textsuperscript{2}Hubei Key Laboratory of Water System Science for Sponge City Construction (Wuhan University), Wuhan 430072, China}\\[0.6ex]
{\footnotesize\textsuperscript{*}Corresponding author.}
\end{minipage}%
}

\begin{document}
\maketitle

\begin{abstract}
Time series forecasters that incorporate exogenous covariates are fragile in deployment: when those covariates are corrupted --- noised, temporally misaligned, or missing --- state-of-the-art exogenous-fusion and exogenous-adapted models (TimeXer, DAG, iTransformer, CrossLinear) degrade 32--62\% above the endogenous-only floor on electricity-price forecasting. A growing line of work responds with increasingly elaborate architectures that bound, gate, or mask the exogenous pathway. We show that much of this architectural advantage can be matched or exceeded by a one-line, model-agnostic training intervention: \textbf{exogenous dropout}, which randomly zeros whole exogenous channels during training. It restores robustness across every architecture we test at negligible cost to clean accuracy; applied to an existing dual-correlation network (DAG), it yields the most robust model in our study --- Gaussian degradation $+$3\%, misalignment $-$8\%, missing-channel $-$4\% relative to the floor, at near-best clean accuracy (MSE 0.259 vs.\ the study-best 0.255). To test whether such robustness instead demands a dedicated \textit{architecture}, we build a deliberately strong bounded foil, \textbf{BoundEx} --- a learnable gate, a fallback residual that provably reverts to the endogenous baseline, and per-channel exogenous FiLM modulation --- and find it outperformed by DAG trained with the same dropout, on clean accuracy and on all three corruption types. An architecture$\times$dropout ablation, a gate-behavior diagnostic, and a representation-level bound (which holds but is empirically loose) show that architectural boundedness is not necessary for this robustness: although the bound does contribute within BoundEx in some domains, an unbounded model trained with the same dropout is more robust than the bounded one in every domain. The finding replicates across three domains (electricity prices, reservoir hydrology, and meteorology) and three corruption types (Gaussian noise, temporal misalignment, and fully missing channels). We release a corruption-robustness benchmark and recommend exogenous dropout as a simple, strong baseline against which future architectural proposals should be measured.
\end{abstract}

\section{Introduction}

\label{sec:introduction}

Electricity price forecasting, weather-driven energy prediction, and supply-chain demand estimation share a common structure: forecasters observe both an endogenous target history and a stream of exogenous covariates --- day-ahead auction prices, temperature forecasts, holiday calendars --- that are known in advance over the forecast horizon. Incorporating these future-known covariates should, in principle, improve accuracy. Yet the dominant paradigm for doing so treats exogenous variables as unconstrained information sources whose influence on the model's internal representation has no architectural upper bound.

This absence of a bound matters profoundly in deployment. Real-world exogenous feeds are routinely corrupted: day-ahead prices are revised after settlement, weather forecasts drift, sensor readings fail, and pipeline errors introduce spurious channels. Under such conditions, unbounded exogenous fusion becomes a liability: when the covariates are corrupted, the exogenous pathway, rather than helping, can actively harm prediction, and no architectural mechanism prevents it from doing so.

This vulnerability is not theoretical, and the field has noticed. A rapidly growing line of work --- TimeXer \cite{wang2024timexer}, ExoTST \cite{tayal2024exotst}, DAG \cite{qiu2025dual}, CrossLinear \cite{zhou2025crosslinear}, CITRAS \cite{yamaguchi2026citras}, and FTimeXer \cite{li2026ftimexer} --- responds with increasingly sophisticated fusion mechanisms (cross-attention, dual temporal-channel correlation graphs, stochastic masking, frequency-domain gating), several explicitly motivated by robustness to imperfect covariates. The implicit premise is that defending the exogenous pathway requires dedicated \textit{architectural} machinery. In this paper we question that premise, and find that a far simpler intervention suffices.

We take a deliberately minimal path: we study a one-line, model-agnostic training intervention --- \textbf{exogenous dropout} --- which randomly zeros entire exogenous channels during training (rescaling the survivors), exposing the model to varying exogenous availability so that it learns not to over-rely on any single covariate. We add it to five representative forecasters and measure robustness under a controlled corruption benchmark. To put the architectural hypothesis to a fair test rather than simply dismiss it, we also build a deliberately strong bounded architecture --- \textbf{BoundEx (Bounded Exogenous Modulation)} --- not as a proposed model, but as a controlled \textit{foil}: it injects exogenous information through a FiLM-style affine modulation \cite{perez2018film} --- channel-wise scaling $\gamma$ and shifting $\beta$ --- gated by a learned scalar $\bar{g} \in [0,1]$ and combined with the endogenous representation through a fallback residual:

\begin{equation}
h_{\text{out}} = (1 - \bar{g}) \odot h_{\text{endog}} + \bar{g} \odot (\gamma \odot h_{\text{endog}} + \beta)
\end{equation}

When the gate closes ($\bar{g} \to 0$) the representation provably reverts to the endogenous backbone (we formalize this bound in Appendix~A), and future-known covariates, when available, enter only through this same bounded modulation pathway. Our central finding is negative for this architectural route: once exogenous dropout is applied to the baselines, BoundEx --- despite its built-in guarantee --- is outperformed on clean accuracy \textit{and} on all three corruption types, most decisively by an existing dual-correlation network (DAG) trained with the same dropout. The robustness reported for such bounded designs, we conclude, comes from the augmentation rather than the architecture.

\textbf{Our contributions} are fourfold:

\begin{enumerate}
  \item We introduce a systematic \textbf{exogenous-corruption robustness benchmark} spanning three domains (electricity prices across five markets, reservoir hydrology, and meteorology), four test-time regimes (clean, Gaussian noise, temporal misalignment, and missing channels), and five random seeds per configuration, with degradation measured against the endogenous-only floor.
  \item We show that \textbf{exogenous dropout}, a one-line training-time augmentation, is a strong and \textit{model-agnostic} robustness baseline: it improves robustness under Gaussian noise, misalignment, and missing channels for every architecture tested (e.g.\ DAG $+$32\%$\to+$3\% Gaussian, $+$25\%$\to-$8\% misalignment, $+$24\%$\to-$4\% missing) at negligible clean-accuracy cost.
  \item Through a complete \textbf{architecture$\times$dropout ablation}, we establish that architectural boundedness is not necessary for this robustness: \textbf{DAG with exogenous dropout is the most robust model in our study while retaining near-best clean accuracy}, and outperforms our purpose-built bounded architecture \textbf{BoundEx} on clean accuracy and on all three corruption types --- in every domain.
  \item We provide a mechanism analysis --- a gate-behavior diagnostic showing the learned gate does \textit{not} close under corruption in any domain, and a representation-level bound that holds but is empirically loose --- supporting the conclusion that architectural boundedness is \textbf{not required} for exogenous-corruption robustness (it does aid BoundEx in some domains, but an unbounded model with the same dropout is more robust everywhere).
\end{enumerate}

\section{Related Work}

\label{sec:related_work}

\subsection{Endogenous Time Series Forecasting}

\label{sec:endogenous_time_series_forecasting}

The past five years have witnessed rapid evolution in deep learning for time series forecasting. Early Transformer-based forecasters adapted the canonical encoder-decoder architecture \cite{vaswani2017attention} to temporal data: Informer \cite{zhou2021informer} introduced sparse attention for long sequences, Autoformer \cite{wu2021autoformer} leveraged seasonal-trend decomposition with auto-correlation, and FEDformer \cite{zhou2022fedformer} operated in the frequency domain. A critical inflection point came with the observation that simpler architectures often outperform complex Transformers on standard benchmarks. DLinear \cite{zeng2023transformers} demonstrated that a single linear layer could surpass many Transformer variants, prompting a re-examination of architectural complexity. PatchTST \cite{nie2022time} addressed this by segmenting time series into patches and applying channel-independent attention, achieving strong results while maintaining Transformer expressivity. More recent work has explored MLP-based architectures \cite{ekambaram2023tsmixer, das2023longterm} and foundation model approaches \cite{ansari2024chronos, liang2024foundation}.

This literature establishes a strong history-only baseline. By improving representations of trend, seasonality, long-range dependence, channel structure, and distributional shift within the lookback window, endogenous forecasters show that substantial predictive power can be extracted without external covariate streams. This strength defines the \textit{endogenous floor}: the accuracy attainable when exogenous information is ignored. Exogenous methods should be judged relative to this floor, improving clean accuracy without falling substantially below it when covariates are noisy, misaligned, missing, or otherwise corrupted.

At the same time, endogenous forecasting leaves a distinct problem unresolved. Many real tasks provide auxiliary signals that differ from target history: they may come from separate systems, be available over the forecast horizon, or be corrupted independently of the endogenous series. The question is therefore not only how to build a stronger history encoder, but how to incorporate such covariates without letting them dominate forecasts when their reliability deteriorates. This motivates the exogenous-variable forecasting literature reviewed next.

\subsection{Exogenous Variables in Time Series Forecasting}

\label{sec:exogenous_variables_in_time_series_forec}

The integration of exogenous variables into neural forecasters has attracted growing attention, driven by applications in energy markets, transportation, and healthcare where future-known covariates are routinely available. Early and general-purpose approaches incorporated covariates directly into the forecasting model. N-BEATSx \cite{olivares2021neural} extended the N-BEATS basis expansion architecture with exogenous stacks, while Temporal Fusion Transformer \cite{lim2021temporal} used variable selection networks to weight covariates before feeding them into a multi-head attention encoder, and iTransformer \cite{liu2023itransformer} inverted the attention dimension --- attending across variates rather than time steps --- so that auxiliary covariates can be handled as part of the multivariate input. These methods, while effective, treat exogenous and endogenous information largely symmetrically --- providing no protection against corrupted covariates.

More recent work has developed specialized architectures for exogenous fusion, in rapid succession. TimeXer \cite{wang2024timexer} introduces a dual-branch Transformer in which endogenous patches and exogenous variables exchange information through cross-attention with learnable global tokens. ExoTST \cite{tayal2024exotst} separates past and future exogenous variables as distinct modalities and fuses them through cross-temporal attention. CATS \cite{lu2024cats} constructs auxiliary time series from the covariates as a plug-and-play exogenous representation. DAG \cite{qiu2025dual} builds dual temporal and channel-wise correlation graphs to discover and inject exogenous dependencies, achieving strong clean-data accuracy. CrossLinear \cite{zhou2025crosslinear} adds a lightweight plug-and-play cross-correlation embedding. The two most recent entries, CITRAS \cite{yamaguchi2026citras} and FTimeXer \cite{li2026ftimexer}, respectively propose a covariate-informed Transformer with dedicated exogenous embedding and fusion layers, and a frequency-domain gated fusion combined with \textit{stochastic exogenous masking} and consistency regularization.

Two properties of this literature are worth emphasizing for what follows. First, to our knowledge none of these methods imposes a \textit{structural} constraint that bounds how much the exogenous pathway can perturb the endogenous representation --- their fusion magnitude is unconstrained by design, so a corrupted covariate can in principle move the forecast arbitrarily far from the endogenous-only prediction. Second, where robustness to imperfect covariates is addressed at all, it is addressed \textit{empirically} rather than architecturally: most tellingly, the robustness component of the most recent of these models, FTimeXer, is closely related to \textit{stochastic exogenous masking} --- a relative of the exogenous dropout we isolate and study in this paper --- rather than to any architectural bound. These two observations --- the absence of an architectural bound, and the empirical, masking-based nature of existing robustness mechanisms --- motivate the two questions this paper asks: whether an explicit architectural bound is necessary, and how far a simple masking-style training augmentation alone can carry robustness.

\subsection{Conditional Modulation and Gating in Deep Learning}

\label{sec:conditional_modulation_and_gating_in_dee}

A long line of work studies how one signal can conditionally modulate another inside a neural network. Feature-wise Linear Modulation (FiLM) \cite{perez2018film} was introduced for visual question answering, where linguistic representations conditionally scale and shift visual features through channel-wise affine parameters; it has since been adopted across image generation, style transfer, and speech processing as a lightweight way to inject conditioning information without restructuring the backbone. Gating mechanisms have an equally long history: LSTMs \cite{hochreiter1997lstm} use forget and input gates to regulate information flow through time; Highway Networks \cite{srivastava2015highway} route signals through gated skip connections that let very deep networks preserve identity mappings; and Mixture-of-Experts architectures \cite{shazeer2017moe} employ learned gating functions to sparsely route inputs to specialized sub-networks. Gated Linear Units \cite{dauphin2017glu} offer a simpler element-wise variant in which a sigmoid gate multiplicatively controls feature propagation. A common principle recurs across these methods --- \textit{conditional, gated information flow}, in which a learned gate determines how strongly an auxiliary or transformed signal influences the main representation and, in the limit of a closed gate, recovers the unmodulated signal.

A complementary ingredient is normalization for distribution shift. Reversible instance normalization (RevIN) \cite{kim2022revin} normalizes each input series and later denormalizes the forecast, improving robustness to shifts in level and scale; a forecaster equipped with instance normalization thus provides a stable, well-characterized baseline that a gated mechanism can fall back to when its gate closes. How these ingredients --- affine FiLM modulation, a learned gate, and a fallback to a normalized endogenous baseline --- can be assembled into an exogenous-injection mechanism with an explicit bound on exogenous influence is the design we study in \S3.

\section{Methods: A Bounded Foil (BoundEx) and Exogenous Dropout}

\label{sec:method}

This section defines the two methods we compare. The intervention the paper ultimately recommends is \textbf{exogenous dropout}, a one-line, model-agnostic training augmentation (\S3.5). Against it we set \textbf{BoundEx}, a deliberately strong \emph{bounded foil}: an architecture purpose-built to make exogenous influence controllable by construction, which we use to test whether architectural boundedness is necessary for corruption robustness (\S3.1--\S3.4). We describe both below so that the comparison in \S4--\S5 is concrete.

\subsection{Problem Formulation}

\label{sec:problem_formulation}

A forecasting model receives three inputs: an endogenous lookback window $X_{\text{endog}} \in \mathbb{R}^{L \times d_e}$ of the target series over $L$ historical time steps with $d_e$ variates; historical exogenous covariates $X_{\text{hist}} \in \mathbb{R}^{L \times d_x}$ observed over the same lookback; and future-known exogenous covariates $X_{\text{future}} \in \mathbb{R}^{H \times d_x}$ spanning the $H$-step forecast horizon. The target is $Y \in \mathbb{R}^{H \times d_e}$. A corruption process $\tilde{X} = \text{Corrupt}(X; c, \sigma)$ maps clean exogenous inputs to degraded ones under corruption type $c$ (Gaussian noise, temporal misalignment, or missing channels) with severity $\sigma$.

The standard approach fuses these inputs through an unbounded mechanism --- cross-attention in TimeXer \cite{wang2024timexer}, dual temporal-channel correlation in DAG \cite{qiu2025dual}, or concatenation followed by a Transformer encoder --- producing a prediction $\hat{y}_{\text{unbounded}} = \text{Fuse}(h_{\text{endog}}, X_{\text{hist}}, X_{\text{future}})$ where $h_{\text{endog}} = \text{Backbone}(X_{\text{endog}})$ is the endogenous encoder output. The key vulnerability is that $\text{Fuse}$ lacks any architectural constraint on $\|\hat{y}_{\text{unbounded}} - \hat{y}_{\text{floor}}\|$, where $\hat{y}_{\text{floor}} = \text{Projector}(h_{\text{endog}})$ is the prediction of the endogenous-only model. When exogenous inputs are corrupted, this unbounded perturbation can drive the forecast arbitrarily far from the safe endogenous baseline.

We define two desiderata for robust exogenous injection:

\begin{enumerate}
  \item \textbf{Gate-proportional influence on the representation:} the deviation of the modulated representation from the endogenous one scales linearly with a learned gate $\bar{g} \in [0,1]$ --- by construction $\|h_{\text{out}} - h_{\text{endog}}\| = \bar{g}\,\|h_{\text{mod}} - h_{\text{endog}}\|$ --- so exogenous influence on the internal representation is controlled \textit{in proportion} to the gate rather than admitted without limit.
  \item \textbf{Graceful degradation:} as corruption severity increases, the gate $\bar{g}$ should approach zero, causing the model to revert to the endogenous-only prediction --- the model may fail to \textit{improve} upon the floor but must not fall \textit{below} it.
\end{enumerate}

\subsection{Exogenous Encoding}

\label{sec:asymmetric_exogenous_encoding}

\begin{figure}[H]
\centering
\includegraphics[width=0.72\columnwidth]{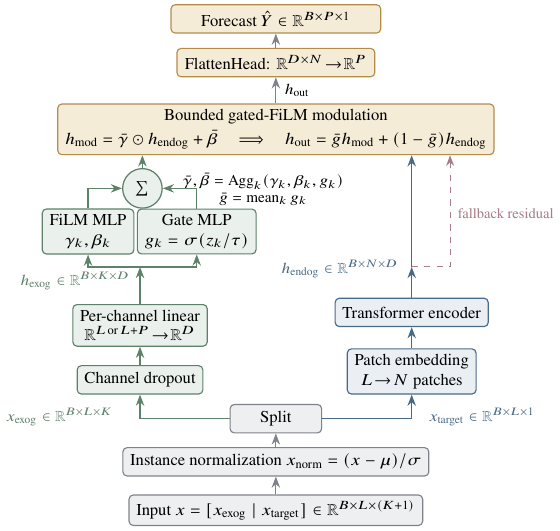}
\caption{Overview of the BoundEx bounded gated-FiLM modulation mechanism. The endogenous PatchTST backbone produces $h_{\text{endog}}$; in parallel, the exogenous channels (optionally concatenated with future-known covariates) pass through channel dropout and the per-channel exogenous encoder to yield the FiLM parameters $\bar{\gamma},\bar{\beta}$ and the scalar gate $\bar{g}$. The bounded modulation $h_{\text{out}} = (1 - \bar{g})\,h_{\text{endog}} + \bar{g}\,(\bar{\gamma} \odot h_{\text{endog}} + \bar{\beta})$ retains a fallback residual to the endogenous backbone (full details in \S3).}
\label{fig:framework_overview}
\end{figure}

BoundEx encodes the exogenous covariates through a lightweight pathway, kept simple so that whatever robustness the architecture exhibits can be attributed to the bounded modulation (\S3.3) rather than to an elaborate encoder. Each of the $K$ exogenous channels is mapped to a single summary vector by a linear projection shared across channels,
\begin{equation}
h_{\text{exo},k} = W_e\, x_{\text{exo},k} + b_e \in \mathbb{R}^{d_m}, \qquad k = 1, \dots, K,
\end{equation}
where $x_{\text{exo},k}$ is the $k$-th channel's covariate series over time.

When future-known covariates are available across the forecast horizon --- such as day-ahead electricity auction prices or weather \textit{forecasts} (as distinct from weather \textit{measurements}) --- each channel's historical and future-known values are concatenated along the time axis before this projection, so that $W_e \in \mathbb{R}^{d_m \times (L+H)}$; otherwise only the $L$ historical steps are used and $W_e \in \mathbb{R}^{d_m \times L}$. The pathway thus yields a set of per-channel exogenous representations $\{h_{\text{exo},k}\}_{k=1}^{K}$, one summary vector per covariate, which drive the bounded modulation below.

\subsection{Bounded FiLM-Style Modulation}

\label{sec:bounded_film_style_modulation}

The core of BoundEx is a gated affine modulation block that limits exogenous influence on the endogenous representation by construction. Given the endogenous representation $h_{\text{endog}} \in \mathbb{R}^{d_m}$ from the PatchTST backbone \cite{nie2022time}, each per-channel exogenous representation $h_{\text{exo},k}$ produces channel-wise FiLM parameters \cite{perez2018film} and a gate:

\begin{equation}
\gamma_k = \text{MLP}_\gamma(h_{\text{exo},k}), \quad \beta_k = \text{MLP}_\beta(h_{\text{exo},k}) \in \mathbb{R}^{d_m}, \quad g_k = \sigma\!\left(\text{MLP}_g(h_{\text{exo},k})/\tau\right) \in [0,1],
\end{equation}

where $\text{MLP}_\gamma, \text{MLP}_\beta$ are two-layer networks initialized near zero (so the modulation starts close to the identity) and $g_k$ is a per-channel scalar gate with temperature $\tau$. The channel parameters are combined into a single FiLM transform and a scalar gate by gate-weighted averaging:

\begin{equation}
\gamma_{\text{agg}} = \frac{\sum_k g_k\,\gamma_k}{\sum_k g_k}, \quad \beta_{\text{agg}} = \frac{\sum_k g_k\,\beta_k}{\sum_k g_k}, \quad \bar{g} = \frac{1}{K}\sum_{k} g_k \in [0,1].
\end{equation}

In practice a small constant $\epsilon = 10^{-8}$ is added to the denominator $\sum_k g_k$ for numerical stability. The modulated representation is then formed as a gated interpolation:

\begin{equation}
\begin{aligned}
h_{\text{mod}} &= \gamma_{\text{agg}} \odot h_{\text{endog}} + \beta_{\text{agg}} \\[4pt]
h_{\text{out}} &= (1 - \bar{g}) \, h_{\text{endog}} + \bar{g} \, h_{\text{mod}}
\end{aligned}
\end{equation}

This formulation provides an immediate architectural guarantee at the representation level. When $\bar{g} = 0$, $h_{\text{out}} = h_{\text{endog}}$ exactly --- the model is indistinguishable from the endogenous-only backbone, regardless of the exogenous input content. When $\bar{g} = 1$, the exogenous modulation operates at full strength. For any intermediate gate value, the representation interpolates linearly between these two extremes. The exogenous contribution to the representation is bounded by:

\begin{equation}
\|h_{\text{out}} - h_{\text{endog}}\| = \bar{g} \cdot \|\gamma_{\text{agg}} \odot h_{\text{endog}} + \beta_{\text{agg}} - h_{\text{endog}}\|,
\end{equation}

which is at most $\bar{g}$ times the maximum modulation magnitude. This is an architectural guarantee on the representation --- it holds for any input, including adversarial perturbations, and does not depend on the training distribution.

We state and prove the corresponding prediction-level bound in Appendix~\ref{app:bound}. In brief, because the forecast head is linear, the forecast's deviation from the gate-closed endogenous floor $\hat{y}_{\text{floor}}$ is controlled \textit{proportionally} by the gate, $\|\hat{y}-\hat{y}_{\text{floor}}\|_2 \le \bar{g}\cdot s\cdot\sigma_{\max}(W)\cdot\|h_{\text{mod}}-h_{\text{endog}}\|_2$, so the prediction collapses to the floor as $\bar{g}\to 0$. This is a \textit{proportional} (not absolute) guarantee --- because the FiLM outputs $\gamma,\beta$ are unconstrained, it is not an absolute, input-independent ball --- and it makes precise what a representation- and prediction-level architectural bound on exogenous influence does and does not provide.

\subsection{Gate Design and Training}

\label{sec:gate_design_and_training}

The overall gate $\bar{g}$ is the mean of the per-channel gates $g_k$, each produced by a compact bottleneck (an MLP with a single hidden layer of width $d_m/2$ followed by a temperature-scaled sigmoid). Averaging over channels makes $\bar{g}$ invariant to the ordering of the exogenous channels --- a useful property when covariates lack a natural ordering or when spurious channels are interleaved with informative ones, since the overall gate depends on the channels' aggregate gating rather than on channel identity or position.

Training minimizes a composite objective:

\begin{equation}
\mathcal{L} = \frac{1}{N H d_e} \sum \|\hat{y} - Y\|_2^2 + \lambda_g \cdot \|\bar{g}\|_1
\end{equation}

The primary term is mean squared error over the forecast horizon. The optional auxiliary term $\lambda_g \cdot \|\bar{g}\|_1$ would encourage gate sparsity, but we find it immaterial: a sweep over $\lambda_g \in \{0, 10^{-4}, 10^{-3}, 10^{-2}\}$ leaves robustness essentially unchanged (Gaussian degradation $+$28.9/$+$28.9/$+$28.5/$+$26.5\% on the dropout-free base). All BoundEx results we report therefore use $\lambda_g = 0$ (pure MSE); the gate is shaped by the data, not by this penalty.

\textbf{Forward pass.} To summarize the computation in one place: the endogenous lookback $X_{\text{endog}}$ is encoded by the PatchTST backbone into $h_{\text{endog}}$; the exogenous covariates (historical, and future-known when available) are encoded by the per-channel linear projection of \S3.2 into the summaries $\{h_{\text{exo},k}\}$; from these the model produces channel-wise FiLM parameters and gates, aggregates them into $\gamma_{\text{agg}}, \beta_{\text{agg}}$ and the scalar gate $\bar{g}$, and combines them with $h_{\text{endog}}$ through the gated residual $h_{\text{out}} = (1-\bar{g})\, h_{\text{endog}} + \bar{g}\,(\gamma_{\text{agg}}\odot h_{\text{endog}}+\beta_{\text{agg}})$ (\S3.3); finally a linear head maps $h_{\text{out}}$ to the forecast $\hat{y}$. The gate $\bar{g}$ is additionally exposed as a diagnostic.

\subsection{Exogenous Dropout}

\label{sec:exogenous_dropout}

The intervention this paper isolates and recommends is a one-line, model-agnostic training augmentation we call \textbf{exogenous dropout}. At each training step, each of the $K$ exogenous channels is independently dropped --- zeroed over its \emph{entire} time span, historical \textit{and} future-known --- with probability $p$, and the surviving channels are rescaled by $1/(1-p)$ so that the expected input is preserved (inverted dropout). Concretely, for an exogenous input $X_{\text{exo}} \in \mathbb{R}^{T \times K}$ (with $T = L$ for historical-only covariates, or $T = L+H$ when future-known values are concatenated), we draw a mask $m \in \{0,1\}^{K}$ with $m_k \sim \text{Bernoulli}(1-p)$ independently per channel and feed
\begin{equation}
\tilde{x}_{\text{exo},k} = \frac{m_k}{1-p}\, x_{\text{exo},k}, \qquad k = 1, \dots, K.
\end{equation}
The mask is \emph{whole-channel} (a covariate is kept or dropped in full, never partially), is \emph{shared} across a channel's historical and future-known segments, is resampled independently at every training step, and is applied \emph{only} during training; at test time all channels pass through unaltered. Exposing the forecaster to a distribution of exogenous-availability patterns teaches it not to over-rely on any single covariate, so that corrupted or absent covariates at test time resemble a regime it has already seen. The augmentation adds no parameters and is agnostic to the backbone; we apply the same rate $p=0.3$ uniformly to \emph{every} model (BoundEx included), so no architecture is privileged by it, and we sweep $p$ in \S4.5.

\section{Experiments}

\label{sec:experiments}

\subsection{Experimental Setup}

\label{sec:experimental_setup}

\textbf{Datasets and preprocessing.} Our corruption-robustness benchmark spans \textbf{three domains}. (i) \textbf{Electricity price forecasting (EPF)} --- five markets, NP (Nord Pool), PJM (Pennsylvania-New Jersey-Maryland), BE (Belgium), FR (France), and DE (Germany) --- each providing hourly electricity prices as the endogenous target and day-ahead auction prices plus auxiliary covariates (load and generation forecasts) as exogenous variables. (ii) \textbf{Reservoir hydrology} --- \textbf{Rapel}, a Chilean hydropower reservoir (daily sampling, 3{,}366 days), forecasting water level from precipitation and inflow. (iii) \textbf{Meteorology} --- \textbf{Weather}, the Jena station (10-minute sampling), forecasting the target variable OT from 20 exogenous meteorological channels. All data are standardized to zero mean and unit variance per channel using training-set statistics and split chronologically into training/validation/test ($\approx$70/10/20). All three domains use a lookback $L = 168$ and forecast horizon $H = 24$ steps --- in each domain's native sampling units (hours for EPF, days for Rapel, 10-minute intervals for Weather) --- and share the same training protocol and model hyperparameters, with no per-domain tuning. The primary metric is Mean Squared Error (MSE), lower is better. Because absolute MSE is not comparable across domains of different scales, we report degradation relative to the endogenous-only floor: for each test-time condition $c \in \{\text{clean, Gaussian, misalign, missing}\}$,
\[
\textbf{$c$ vs-Floor (\%)} = (\text{MSE}_c - \text{MSE}_{\text{floor}}) / \text{MSE}_{\text{floor}} \times 100,
\]
where $\text{MSE}_{\text{floor}}$ is the endogenous-only PatchTST MSE. Negative values indicate performance at or below the floor.

\textbf{Exogenous covariates: historical vs.\ future-known.} Following the standard exogenous-forecasting protocol \cite{wang2024timexer}, every exogenous channel is supplied --- and used by all models --- over both the lookback and the forecast horizon. On EPF this is realistic: day-ahead prices and the load/generation forecasts for the delivery day are published before it begins. On Rapel and Weather we instead supply the recorded future values as an idealized ``perfect-forecast'' stream, as is conventional on these benchmarks; this is conservative for our thesis, since the clean condition hands every model the most informative covariates available, yet exogenous dropout still helps under corruption. All corruption operators act on the full exogenous window (historical \emph{and} future-known); the endogenous target history is never corrupted.

\textbf{Corruption protocol.} We evaluate under four exogenous conditions:

\begin{itemize}
  \item \textbf{Clean:} exogenous variables are used as-is at test time (no corruption).
  \item \textbf{Gaussian:} independent Gaussian noise $\mathcal{N}(0, \sigma^2)$ is added to every exogenous channel, where $\sigma$ is set to the per-channel empirical standard deviation estimated from the training set, producing a signal-to-noise ratio of approximately 1.
  \item \textbf{Misalign:} the exogenous time series is shifted in time relative to the endogenous series by a random offset of $k$ steps ($k \sim \text{Uniform}\{3, \dots, \min(24, T)\}$), simulating the practical scenario where day-ahead price forecasts are revised after settlement.
  \item \textbf{Missing:} all exogenous channels are set to zero at test time, simulating a complete loss of the covariate feed. This applies the same zeroing operation as the train-time exogenous-dropout intervention (\S3.5), but at a more extreme, all-channel level: training drops a random subset of channels, whereas this condition removes all of them at once.
\end{itemize}

\textbf{Exogenous dropout (the studied intervention).} We apply the exogenous-dropout augmentation defined in \S3.5 (whole-channel, $p=0.3$, inverted scaling, train-only) to \textit{every} architecture, both with and without it --- matched across models --- yielding the architecture$\times$dropout comparison in \S4.2.

\textbf{Baselines.} We compare against five models:

\begin{center}
\resizebox{\columnwidth}{!}{%
\begin{tabular}{ll}
\toprule
\textbf{Model} & \textbf{Description} \\
\midrule
\textbf{PatchTST} \cite{nie2022time} & Endogenous-only baseline; the \textit{endogenous floor} --- performance achievable when exogenous variables are ignored entirely. \\
\textbf{TimeXer} \cite{wang2024timexer} & Symmetric cross-attention between endogenous patches and exogenous tokens via learnable global query tokens. \\
\textbf{DAG} \cite{qiu2025dual} & Dual temporal and channel correlation graphs to discover and inject exogenous dependencies. \\
\textbf{iTransformer} \cite{liu2023itransformer} & Inverts the standard Transformer to attend across variates; exogenous channels provided as additional variates. \\
\textbf{CrossLinear} \cite{zhou2025crosslinear} & Plug-and-play cross-correlation embedding that linearly aligns exogenous with endogenous series before a shared forecasting head. \\
\bottomrule
\end{tabular}
}
\end{center}

All baselines are trained with their authors' recommended configurations. An additional configuration, \textbf{BoundEx (no fallback)}, is our method with the $(1 - \bar{g})$ residual connection removed --- the exogenous modulation output $h_{\text{mod}}$ replaces $h_{\text{endog}}$ directly rather than interpolating --- serving as an ablation that isolates the effect of boundedness. We do not include the following models as baselines: ExoTST \cite{tayal2024exotst}, CITRAS \cite{yamaguchi2026citras}, and FTimeXer \cite{li2026ftimexer} have no official public code; CATS \cite{lu2024cats} is available but needs substantial per-dataset re-tuning to run faithfully under our shared, no-per-dataset-tuning protocol; and N-BEATSx \cite{olivares2021neural} and the Temporal Fusion Transformer \cite{lim2021temporal} are earlier models we do not consider here.

\textbf{Training and inference.} All models are trained for 50 epochs with early stopping (patience = 5) using the Adam optimizer with learning rate $10^{-4}$ and MSE loss, with five random seeds (0--4) per dataset. On EPF this yields 25 seed-runs per model across the five markets, from which we report per-market test-set MSE and cross-market means with standard error across markets ($\text{SEM} = \text{SD} / \sqrt{5}$, treating markets as the independent clustering unit for a conservative uncertainty estimate); on Rapel and Weather we report the five-seed mean. All experiments run on a single NVIDIA GeForce RTX 5070 Laptop GPU (8 GB VRAM) with PyTorch 2.10 and CUDA 13.0.

\textbf{Hyperparameters.} Table 1 summarizes the shared and model-specific hyperparameters.

\begin{table}[H]
\centering
\caption{Training protocol. Shared settings apply to all models; baselines otherwise use their authors'/TFB configurations (e.g.\ TimeXer $d_{\text{model}}{=}256$, DAG $d_{\text{model}}{=}64$), not a forced common size.}
\label{tab:2}
\begin{tabular}{ll}
\toprule
\textbf{Setting} & \textbf{Value} \\
\midrule
Lookback $L$ / Horizon $H$ & 168 / 24 \\
Epochs / Patience & 50 / 5 \\
Optimizer & Adam \\
Exogenous dropout (studied factor) & $\{0,\ 0.3\}$ --- whole-channel, inverted, train-only \\
\midrule
BoundEx: $d_{\text{model}}$ / layers / heads & 128 / 2 / 8 \\
BoundEx: patch len / stride & 16 / 8 \\
BoundEx: gate reg.\ $\lambda_g$ & 0 (immaterial; see \S3.4) \\
\bottomrule
\end{tabular}
\end{table}

\subsection{Main Results}

\label{sec:main_results}

Table 2 reports our main results: the architecture $\times$ exogenous-dropout comparison across all three domains (five EPF markets, Rapel, and Weather), expressed as vs-Floor degradation under each test-time condition. It answers the core question of this paper: under corrupted --- or entirely missing --- exogenous inputs, does a model stay close to the safe endogenous-only floor, or degrade far above it?

\begin{table}[H]
\centering
\caption{Main results: architecture $\times$ exogenous-dropout across all three domains, as vs-Floor degradation (\%) under each test-time condition (lower is better; negative beats the endogenous-only floor). \textbf{Bold} marks the best (lowest) value in each column within each domain block; \underline{underline} marks the second-best.}
\label{tab:main}
\small
\begin{tabular}{llrrrr}
\toprule
\textbf{Model} & \textbf{Exog-drop} & \textbf{Clean} & \textbf{Gaussian} & \textbf{Misalign} & \textbf{Missing} \\
\midrule
\multicolumn{6}{l}{\textit{Electricity prices (EPF; 5 markets, hourly)}} \\
iTransformer & no  & $+$15.5 & $+$61.9 & $+$60.8 & $+$51.9 \\
iTransformer & yes & $+$9.9  & $+$31.4 & $+$30.5 & $+$35.2 \\
TimeXer      & no  & $-$17.9 & $+$34.3 & $+$19.0 & $+$27.5 \\
TimeXer      & yes & $-$17.3 & $+$25.1 & $+$7.1  & $+$13.2 \\
DAG          & no  & \textbf{$-$27.5} & $+$32.2 & $+$24.8 & $+$23.7 \\
DAG          & yes & \underline{$-$26.1} & \textbf{$+$3.4} & \textbf{$-$7.9} & \textbf{$-$4.3} \\
CrossLinear  & no  & $+$15.5 & $+$62.3 & $+$48.5 & $+$50.7 \\
CrossLinear  & yes & $+$16.8 & $+$32.3 & $+$24.2 & $+$29.1 \\
BoundEx (ours) & no  & $-$16.5 & $+$28.9 & $+$1.3 & $+$28.3 \\
BoundEx (ours) & yes & $-$11.9 & \underline{$+$11.3} & \underline{$-$1.4} & \underline{$+$3.6} \\
\midrule
\multicolumn{6}{l}{\textit{Reservoir hydrology (Rapel; daily, target $=$ water level)}} \\
iTransformer & no  & $+$18.3 & $+$38.2 & $+$30.7 & $+$22.1 \\
iTransformer & yes & $+$19.5 & $+$36.1 & $+$29.8 & $+$19.3 \\
TimeXer      & no  & $-$21.1 & $+$15.7 & $+$7.8  & $+$14.2 \\
TimeXer      & yes & \textbf{$-$27.1} & $+$5.0  & $-$2.8  & $+$2.1 \\
DAG          & no  & $-$19.7 & \underline{$-$13.4} & \underline{$-$15.4} & \underline{$-$8.4} \\
DAG          & yes & \underline{$-$21.5} & \textbf{$-$16.8} & \textbf{$-$18.4} & \textbf{$-$11.2} \\
CrossLinear  & no  & $+$30.3 & $+$45.8 & $+$38.9 & $+$20.9 \\
CrossLinear  & yes & $+$22.3 & $+$38.1 & $+$25.7 & $+$22.8 \\
BoundEx (ours) & no  & $-$8.6 & $+$2.1 & $-$10.1 & $-$2.8 \\
BoundEx (ours) & yes & $-$7.4 & $+$3.0 & $-$8.2 & $-$3.5 \\
\midrule
\multicolumn{6}{l}{\textit{Meteorology (Weather; 10-min, 20 exogenous channels)}} \\
iTransformer & no  & $+$86.5  & $+$526.7 & $+$107.1 & $+$492.7 \\
iTransformer & yes & $+$119.5 & $+$445.4 & $+$138.2 & $+$401.7 \\
TimeXer      & no  & $+$9.3 & $+$45.9 & $+$31.9 & $+$42.7 \\
TimeXer      & yes & $+$9.1 & $+$41.6 & $+$31.4 & $+$36.4 \\
DAG          & no  & \textbf{$-$43.8} & \underline{$-$10.3} & \textbf{$-$19.3} & \underline{$-$14.5} \\
DAG          & yes & \underline{$-$39.2} & \textbf{$-$12.3} & \underline{$-$12.8} & \textbf{$-$15.3} \\
CrossLinear  & no  & $+$133.8 & $+$484.3 & $+$144.0 & $+$461.5 \\
CrossLinear  & yes & $+$106.0 & $+$306.8 & $+$115.8 & $+$318.6 \\
BoundEx (ours) & no  & $-$8.2 & $+$26.4 & $+$2.8 & $+$38.8 \\
BoundEx (ours) & yes & $-$3.9 & $-$0.8 & $+$3.7 & $-$3.5 \\
\bottomrule
\end{tabular}
\end{table}

\noindent\textit{``Exog-drop'' is whether exogenous-channel dropout ($p=0.3$) is applied during training. CrossLinear (all domains) and iTransformer on Weather do \emph{not} train stably under our shared, no-per-dataset-tuning protocol; we report their true numbers rather than omit them (training and stability details, including the per-baseline divergence counts, are in Appendix~\ref{app:baselines}).}

The table establishes the paper's central empirical claims:

\textbf{Exogenous dropout helps across architectures and domains.} On EPF, adding the augmentation lowers Gaussian degradation for all five models --- iTransformer $+$62$\to+$31, TimeXer $+$34$\to+$25, DAG $+$32$\to+$3, CrossLinear $+$62$\to+$32, BoundEx $+$29$\to+$11 --- and likewise under misalignment and missing channels, while clean accuracy barely moves (absolute MSE, e.g.\ DAG 0.255$\to$0.259). The effect is not specific to electricity: on Weather, TimeXer's Gaussian degradation falls $+$45.9$\to+$41.6 and missing $+$42.7$\to+$36.4; on Rapel, TimeXer Gaussian $+$15.7$\to+$5.0. The single most dramatic case is our own bounded BoundEx on Weather, which collapses under corruption without the augmentation (Gaussian $+$26.4, Missing $+$38.8) yet becomes robust with it (Gaussian $-$0.8, Missing $-$3.5). Missing channels --- the same zeroing operation as the augmentation, taken to the all-channel extreme --- behave the same way throughout: having been trained with whole channels randomly zeroed, the dropout models treat a fully-absent covariate feed as in-distribution (on EPF, DAG $+$23.7$\to-$4.3; BoundEx $+$28.3$\to+$3.6). The intervention is one line of code, model-agnostic, and effectively free.

\textbf{The simple augmentation beats the purpose-built architecture, in every domain.} Once the baselines receive the same dropout, BoundEx's apparent advantage disappears. On EPF, DAG$+$dropout attains near-best clean accuracy (absolute MSE 0.259, second only to its own dropout-free variant at 0.255), the best Gaussian ($+$3.4 vs.\ BoundEx's $+$11.3), misalignment ($-$7.9 vs.\ $-$1.4), and missing-channel ($-$4.3 vs.\ $+$3.6) robustness, dominating BoundEx on every axis; across the 25 matched (market, seed) cells this dominance is significant under every condition --- clean $-$14.2, Gaussian $-$7.9, misalignment $-$6.5, and missing $-$7.9 percentage points of vs-Floor degradation in DAG's favor (paired $t$-test and Wilcoxon signed-rank both $p<0.001$; 95\% bootstrap confidence intervals exclude zero). The same holds in the other domains: DAG$+$dropout is the most corruption-robust configuration on Rapel (best in all three corruption columns: Gaussian $-$16.8, Misalign $-$18.4, Missing $-$11.2) and on Weather (best on Gaussian and Missing; on misalignment the already-robust DAG benefits little from dropout). Across the five seeds on Rapel and Weather, DAG$+$dropout has the lower vs-Floor degradation than BoundEx$+$dropout in four to five of the five seeds under every corruption. BoundEx$+$dropout remains competitive --- ahead of the dropout-augmented iTransformer, TimeXer, and CrossLinear --- but is dominated by DAG, so the bounded design confers no advantage in any domain.

\textbf{Boundedness is not necessary for robustness.} On EPF, stripping the exogenous dropout from BoundEx (otherwise identical) inflates Gaussian degradation from $+$11\% to $+$29\%, back to the level of the unbounded baselines, even though the bounded gate and fallback residual are untouched, while adding dropout to an unbounded model (DAG) drops it to $+$3\%. The two-factor ablation of \S4.4 shows the precise balance between the bound and the augmentation shifts across domains, but one comparison holds everywhere: an unbounded DAG trained with the same dropout is the most robust model in all three domains, so no architectural bound is required. We analyze the limited role of the learned gate in \S5.

\begin{figure}[H]
\centering
\includegraphics[width=\columnwidth]{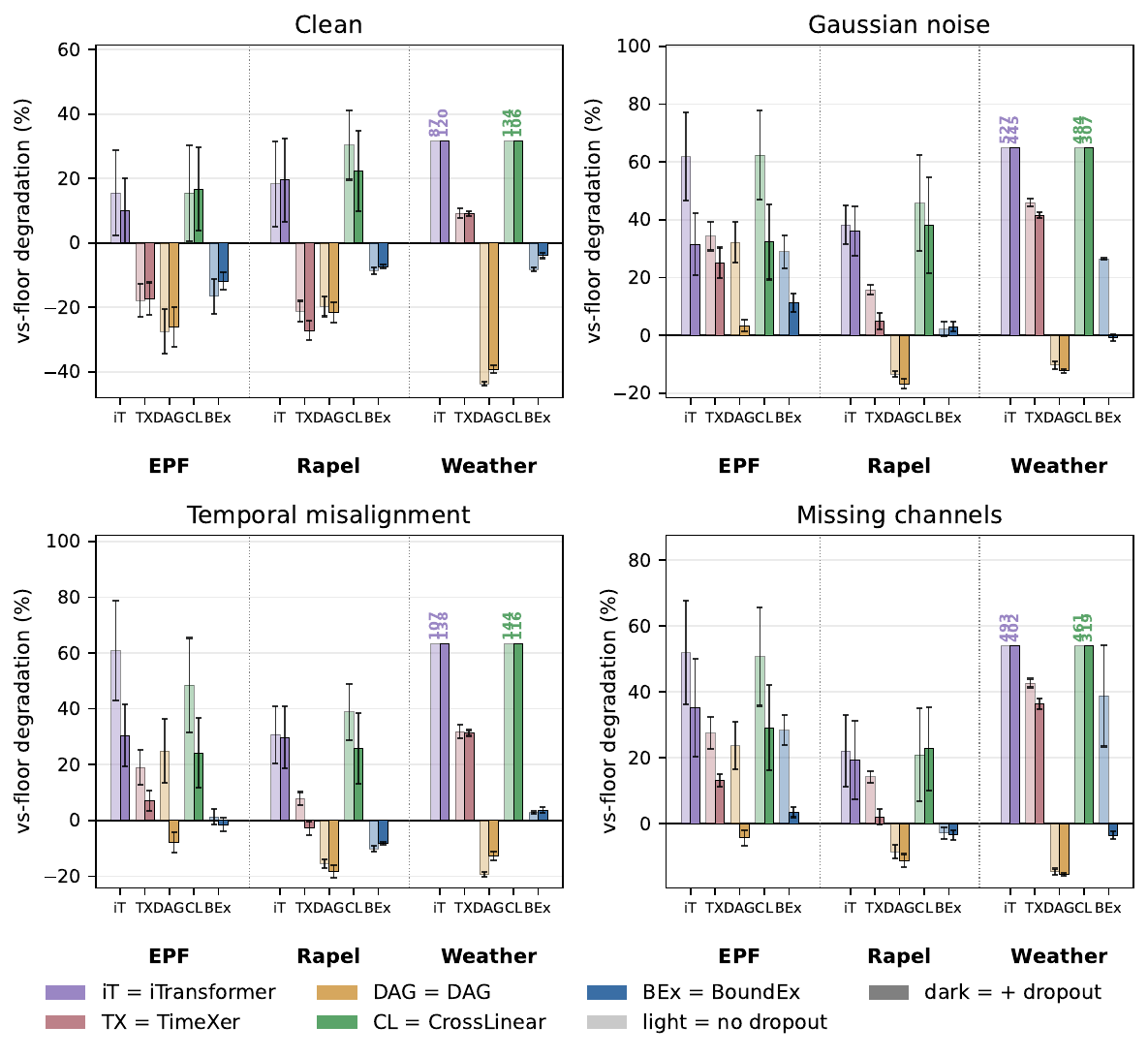}
\caption{Effect of exogenous dropout on vs-Floor degradation (\%; lower is better; light $=$ no dropout, dark $=$ $+$dropout; error bars $\pm$1 SEM), one panel per test-time condition (clean, Gaussian noise, temporal misalignment, missing channels). Each panel groups the five models (iT $=$ iTransformer, TX $=$ TimeXer, DAG, CL $=$ CrossLinear, BEx $=$ BoundEx) within each of the three domains (EPF, Rapel, Weather). The $y$-axis is scaled per panel and bars above the cap are clipped with their true value printed.}
\label{fig:head_to_head_robustness}
\end{figure}

\begin{figure}[H]
\centering
\includegraphics[width=\columnwidth]{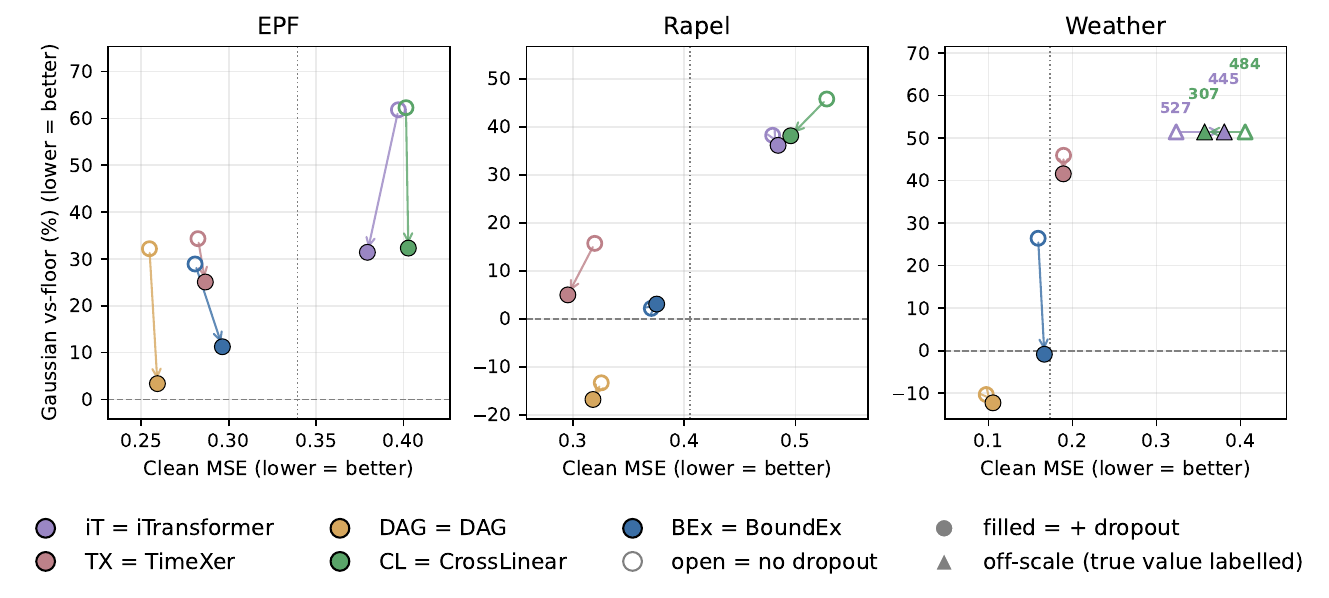}
\caption{Accuracy--robustness trade-off, one panel per domain (absolute clean MSE is comparable only within a domain): clean MSE (x-axis) vs.\ Gaussian degradation above the floor (y-axis), both lower-better; each of the five models appears without (open) and with (filled) exogenous dropout, the arrow showing dropout's effect, and the dotted line marks the endogenous floor. Up-triangles ($\blacktriangle$) mark points that run off the top of the panel, with their true value printed.}
\label{fig:accuracy_robustness_frontier}
\end{figure}

\subsection{Per-Market Head-to-Head}

\label{sec:per_market_consistency}

As a within-EPF robustness check, Table 3 breaks the dropout-matched DAG vs.\ BoundEx comparison (both trained with dropout) down by individual market, confirming that the headline numbers of Table 2 are not a cross-market averaging artifact. DAG$+$dropout has the lower degradation in every market on clean accuracy, in four of five markets under both Gaussian noise and missing channels, and in three of five under misalignment --- the regime where BoundEx's bounded modulation is genuinely competitive. In no market does BoundEx recover the decisive advantage it appeared to hold before the baselines were given the same augmentation; once dropout is shared, the bounded design earns no consistent per-market edge.

\begin{table}[H]
\centering
\caption{Per-market head-to-head, dropout-matched: DAG$+$dropout vs.\ BoundEx$+$dropout (vs-Floor \%, 5 seeds per market; lower is better, negative beats the endogenous floor). Both models are trained with exogenous dropout $p=0.3$; cross-market means appear in Table 2. \textbf{Bold} marks the more robust (lower) of the two models for each market and condition.}
\label{tab:permarket}
\small
\begin{tabular}{lrrrr}
\toprule
\textbf{Market} & \textbf{Clean} & \textbf{Gaussian} & \textbf{Misalign} & \textbf{Missing} \\
\midrule
\multicolumn{5}{l}{\textit{DAG $+$ dropout}} \\
NP  & \textbf{$-$23.7} & \textbf{$-$3.1} & \textbf{$-$12.6} & \textbf{$-$7.7} \\
PJM & \textbf{$-$45.3} & \textbf{$+$5.7} & \textbf{$-$9.7}  & \textbf{$-$12.0} \\
BE  & \textbf{$-$11.2} & $+$7.4 & $+$0.1  & $+$2.8 \\
FR  & \textbf{$-$12.3} & \textbf{$+$8.5} & $+$2.4  & \textbf{$+$0.1} \\
DE  & \textbf{$-$38.0} & \textbf{$-$1.5} & \textbf{$-$19.8} & \textbf{$-$4.5} \\
\midrule
\multicolumn{5}{l}{\textit{BoundEx $+$ dropout}} \\
NP  & $-$8.7  & $+$4.4  & $-$0.4  & $-$0.7 \\
PJM & $-$12.4 & $+$23.9 & $+$6.3  & $+$5.9 \\
BE  & $-$11.1 & \textbf{$+$6.6}  & \textbf{$-$4.6}  & \textbf{$+$1.8} \\
FR  & $-$4.2  & $+$13.6 & \textbf{$+$1.8}  & $+$9.3 \\
DE  & $-$22.9 & $+$7.9  & $-$10.2 & $+$1.7 \\
\bottomrule
\end{tabular}
\end{table}

\subsection{Ablation: Dropout vs.\ Boundedness}

\label{sec:ablation}

To partition the contributions of the architectural bound and the dropout augmentation \textit{within} BoundEx, we run the full two-factor ablation --- fallback residual present/absent $\times$ exogenous dropout on/off --- in all three domains, holding everything else fixed. Table 4 reports clean accuracy and degradation under each of the three corruptions for the four cells.

\begin{table}[H]
\centering
\caption{Two-factor ablation within BoundEx: fallback residual (absent \texttt{no-fb} / present \texttt{fb}) $\times$ exogenous dropout (off/on), as vs-Floor (\%) across three domains and four conditions; lower is better. Within each condition the two rows (dropout off/on) and the column pair (\texttt{no-fb}/\texttt{fb}) form the 2$\times$2 grid, and \textbf{bold} marks its best (lowest) cell. EPF: cross-market means (5 markets $\times$ 5 seeds); Rapel/Weather: 5-seed means.}
\label{tab:5}
\small
\setlength{\tabcolsep}{4pt}
\begin{tabular}{l rr rr rr rr}
\toprule
 & \multicolumn{2}{c}{\textbf{Clean}} & \multicolumn{2}{c}{\textbf{Gaussian}} & \multicolumn{2}{c}{\textbf{Misalign}} & \multicolumn{2}{c}{\textbf{Missing}} \\
\cmidrule(lr){2-3}\cmidrule(lr){4-5}\cmidrule(lr){6-7}\cmidrule(lr){8-9}
 & \texttt{no-fb} & \texttt{fb} & \texttt{no-fb} & \texttt{fb} & \texttt{no-fb} & \texttt{fb} & \texttt{no-fb} & \texttt{fb} \\
\midrule
\multicolumn{9}{l}{\textit{EPF (five electricity markets)}} \\
\quad no dropout  & $-$15.7 & \textbf{$-$16.5} & $+$81.2 & $+$28.9 & $+$34.3 & $+$1.3 & $+$86.7 & $+$28.3 \\
\quad $+$ dropout & $-$12.2 & $-$11.9 & $+$25.8 & \textbf{$+$11.3} & $+$7.3 & \textbf{$-$1.4} & $+$23.8 & \textbf{$+$3.6} \\
\midrule
\multicolumn{9}{l}{\textit{Rapel (reservoir hydrology)}} \\
\quad no dropout  & $+$43.7 & \textbf{$-$8.6} & $+$165.7 & \textbf{$+$2.1} & $+$78.9 & \textbf{$-$10.1} & $+$66.5 & $-$2.8 \\
\quad $+$ dropout & $+$55.1 & $-$7.4 & $+$116.0 & $+$3.0 & $+$77.6 & $-$8.2 & $+$82.8 & \textbf{$-$3.4} \\
\midrule
\multicolumn{9}{l}{\textit{Weather (meteorology)}} \\
\quad no dropout  & $-$0.8 & \textbf{$-$8.2} & $+$68.7 & $+$26.4 & $+$3.5 & \textbf{$+$2.8} & $+$69.9 & $+$38.8 \\
\quad $+$ dropout & $-$0.3 & $-$3.9 & $+$38.4 & \textbf{$-$0.8} & $+$3.7 & $+$3.7 & $+$35.7 & \textbf{$-$3.5} \\
\bottomrule
\end{tabular}
\end{table}

\textbf{The decomposition is domain- and corruption-dependent.} On EPF the two single factors are close, but which one leads depends on the corruption. Reading from the neither-factor cell (no fallback, no dropout), dropout alone is the marginally stronger factor under Gaussian noise ($+$81$\to+$26\%, against $+$29\% for the fallback alone) and missing channels ($+$87$\to+$24\%, against $+$28\%), whereas the fallback alone is the stronger factor under misalignment ($+$34$\to+$1\%, against $+$7\% for dropout); only their combination is best on every corruption (Gaussian $+$11\%, misalignment $-$1\%, missing $+$4\%). On Weather and Rapel the order reverses and the fallback residual becomes the dominant factor across corruptions. On Weather the fallback cuts Gaussian degradation from $+$69\% to $+$26\% (against $+$38\% for dropout alone) and missing from $+$70\% to $+$39\% (with dropout alone reaching a comparable $+$36\%), while misalignment is already benign in every cell ($\le+$4\%). On Rapel the fallback is decisive under \textit{every} condition: removing it inflates Gaussian past $+$160\%, misalignment to $+$79\%, and missing to $+$67\%, and harms even \textit{clean} accuracy, which jumps from $-$8\% (with the fallback) to $+$44$\sim$55\% above the floor with no corruption at all, because BoundEx's endogenous path runs through that residual; there dropout alone barely helps, and on clean and missing it even slightly hurts. In these two exogenous-light domains a bounded BoundEx is thus already robust \textit{without} dropout (Rapel at fallback, no-dropout: Gaussian $+$2\%, misalignment $-$10\%, missing $-$3\%).

\textbf{What generalizes is the comparison that matters for our thesis.} An unbounded model carries no such fallback, yet --- trained with the \textit{same} dropout --- DAG is more corruption-robust than BoundEx in every domain (Table 2; Rapel Gaussian $-$17\% vs.\ BoundEx $+$3\%, Weather $-$12\% vs.\ $-$1\%, EPF $+$3\% vs.\ $+$11\%). Architectural boundedness therefore \textit{helps} BoundEx in some domains, but it is never \textit{necessary} for robustness: a model with no bound at all, plus dropout, does better. The takeaway is not that the bound is useless within BoundEx (on Weather and Rapel it clearly is not) but that it is not what separates the best models --- the transferable dropout augmentation, applied to a stronger backbone, wins regardless.

\subsection{Dropout-Rate Sensitivity}

\label{sec:psweep}

A natural concern is whether our headline --- exogenous dropout beats architecture, with DAG$+$dropout best --- is an artifact of the single rate $p=0.3$ fixed throughout our protocol. We therefore sweep $p \in \{0, 0.1, 0.3, 0.5, 0.7\}$ for the two central models (the winning DAG and the bounded BoundEx) in all three domains, holding everything else fixed (Table 5, Figure 4).

\begin{table}[H]
\centering
\small
\setlength{\tabcolsep}{5pt}
\caption{Exogenous-dropout rate sweep for DAG and BoundEx across all three domains (vs-Floor \%; EPF: cross-market means, 5 markets $\times$ 5 seeds; Rapel/Weather: 5-seed means; lower is better, \textbf{bold} $=$ per-column best for each model (DAG and BoundEx) within each domain). $p=0$ and $p=0.3$ reuse earlier runs; $p\in\{0.1,0.5,0.7\}$ are new.}
\label{tab:psweep}
\begin{tabular}{llrrrr}
\toprule
\textbf{Model} & $\boldsymbol{p}$ & \textbf{Clean} & \textbf{Gaussian} & \textbf{Misalign} & \textbf{Missing} \\
\midrule
\multicolumn{6}{l}{\textit{EPF (five electricity markets)}} \\
DAG     & 0.0 & \textbf{$-$27.5} & $+$32.2 & $+$24.8 & $+$23.7 \\
DAG     & 0.1 & $-$27.5 & $+$17.3 & $+$7.3  & $+$8.6 \\
DAG     & 0.3 & $-$26.1 & $+$3.4  & $-$7.9  & $-$4.3 \\
DAG     & 0.5 & $-$24.2 & $-$2.8  & $-$12.6 & $-$8.8 \\
DAG     & 0.7 & $-$21.9 & \textbf{$-$6.2} & \textbf{$-$13.9} & \textbf{$-$10.4} \\
BoundEx & 0.0 & \textbf{$-$16.5} & $+$28.9 & $+$1.3  & $+$28.3 \\
BoundEx & 0.1 & $-$15.1 & $+$17.1 & $-$1.2  & $+$9.7 \\
BoundEx & 0.3 & $-$11.9 & $+$11.3 & \textbf{$-$1.4}  & $+$3.6 \\
BoundEx & 0.5 & $-$3.8  & \textbf{$+$11.0} & $+$3.1  & \textbf{$+$3.5} \\
BoundEx & 0.7 & $+$11.7 & $+$13.1 & $+$15.2 & $+$4.0 \\
\midrule
\multicolumn{6}{l}{\textit{Rapel (reservoir hydrology)}} \\
DAG     & 0.0 & $-$19.7 & $-$13.4 & $-$15.4 & $-$8.4 \\
DAG     & 0.1 & $-$20.5 & $-$15.7 & $-$16.8 & $-$10.5 \\
DAG     & 0.3 & \textbf{$-$21.5} & $-$16.8 & \textbf{$-$18.4} & $-$11.2 \\
DAG     & 0.5 & $-$19.8 & \textbf{$-$17.1} & $-$17.4 & \textbf{$-$12.3} \\
DAG     & 0.7 & $-$17.3 & $-$15.8 & $-$15.9 & $-$11.5 \\
BoundEx & 0.0 & \textbf{$-$8.6}  & \textbf{$+$2.1}  & \textbf{$-$10.1} & $-$2.8 \\
BoundEx & 0.1 & $-$8.4  & $+$2.8  & $-$9.7  & $-$3.4 \\
BoundEx & 0.3 & $-$7.4  & $+$3.0  & $-$8.2  & $-$3.4 \\
BoundEx & 0.5 & $-$4.4  & $+$3.8  & $-$5.4  & \textbf{$-$4.6} \\
BoundEx & 0.7 & $-$1.6  & $+$6.7  & $-$1.3  & $-$4.1 \\
\midrule
\multicolumn{6}{l}{\textit{Weather (meteorology)}} \\
DAG     & 0.0 & \textbf{$-$43.8} & $-$10.3 & \textbf{$-$19.3} & $-$14.5 \\
DAG     & 0.1 & $-$42.4 & $-$11.8 & $-$17.9 & $-$14.3 \\
DAG     & 0.3 & $-$39.2 & \textbf{$-$12.3} & $-$12.8 & \textbf{$-$15.3} \\
DAG     & 0.5 & $-$34.8 & $-$11.3 & $-$9.2  & $-$15.1 \\
DAG     & 0.7 & $-$31.3 & $-$8.0  & $-$5.1  & $-$13.7 \\
BoundEx & 0.0 & \textbf{$-$8.2}  & $+$26.4 & \textbf{$+$2.8}  & $+$38.8 \\
BoundEx & 0.1 & $-$6.2  & $+$4.6  & $+$3.6  & $+$4.5 \\
BoundEx & 0.3 & $-$3.9  & \textbf{$-$0.8}  & $+$3.7  & \textbf{$-$3.5} \\
BoundEx & 0.5 & $-$2.3  & $-$0.3  & $+$3.5  & $-$3.5 \\
BoundEx & 0.7 & $+$1.8  & $+$0.9  & $+$6.5  & $-$3.4 \\
\bottomrule
\end{tabular}
\end{table}

\begin{figure}[H]
\centering
\makebox[\columnwidth][c]{\includegraphics[width=0.95\columnwidth]{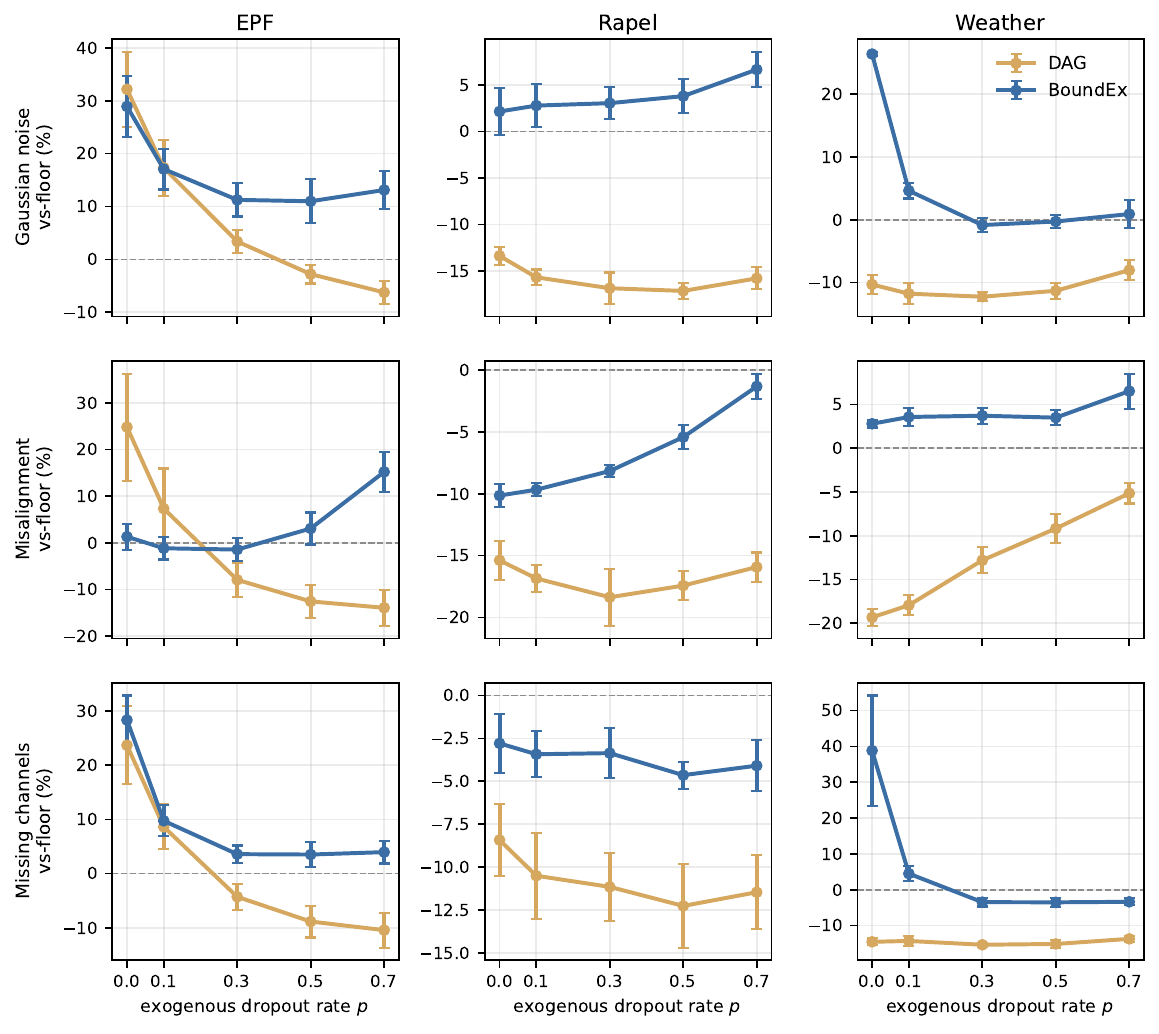}}
\caption{Exogenous-dropout rate ($p$) sensitivity across all three domains (columns) under Gaussian noise (top), misalignment (middle), and missing channels (bottom), for DAG (the winner) and BoundEx (the bounded foil); $y$-axes are vs-Floor degradation and are scaled per panel, with error bars showing $\pm$1 SEM (across the 5 markets on EPF, across the 5 seeds on Rapel and Weather).}
\label{fig:psweep}
\end{figure}

The sweep tells a clean story with a sharp contrast between the two models, and the contrast is consistent across all three domains. For \textbf{DAG}, increasing $p$ reduces degradation under all three corruptions --- on EPF Gaussian $+$32\%$\to-$6\%, misalignment $+$25\%$\to-$14\%, missing $+$24\%$\to-$10\% --- while clean accuracy barely moves (EPF clean vs-Floor drifts only from $-$28\% to $-$22\%, staying well below the floor throughout). The same monotone, rate-forgiving pattern holds on Rapel and Weather, where DAG remains $17$--$44\%$ below its clean floor across the whole range while its corruption degradation falls or holds. There is thus a wide, forgiving operating range in every domain: any $p \in [0.3, 0.7]$ leaves DAG more robust than every architectural baseline at little clean-accuracy cost. The headline is therefore not special to $p=0.3$; if anything, $p=0.3$ is conservative.

\textbf{BoundEx}, by contrast, is markedly more sensitive to the rate. On EPF its robustness has a narrow sweet spot around $p \in [0.1, 0.3]$ and degrades on \textit{every} metric at high $p$: clean accuracy collapses from better-than-floor ($-$17\%) to worse-than-floor ($+$12\%) at $p=0.7$, Gaussian degradation flattens rather than continuing to fall, and misalignment robustness breaks down entirely (from $-$1\% at $p=0.3$ to $+$15\% at $p=0.7$). The purpose-built architecture is thus the more fragile, more hyperparameter-sensitive of the two: heavy exogenous dropout fights its modulation pathway, whereas the same intervention on an unbounded backbone is uniformly beneficial. Most tellingly, in \textit{every} domain and at \textit{every} corruption column, DAG's best rate beats BoundEx's best rate (Table 5, comparing the two models' bold per-column entries): the simple augmentation on a stronger backbone dominates the bounded design across the whole rate axis, not at a single tuned point.

\section{Discussion}

\label{sec:discussion}

The central finding of this study is methodological: the robustness gains commonly attributed to specialized exogenous architectures can be obtained --- and exceeded --- by a one-line training augmentation. Exogenous dropout improves every architecture we test, and an existing unbounded model (DAG) plus dropout is the most robust model in our study --- at near-best clean accuracy --- ahead of the bounded BoundEx we built specifically for robustness.

\textbf{Why does such a simple augmentation work?} Randomly zeroing whole exogenous channels during training exposes the forecaster to a distribution of exogenous-availability patterns, teaching it to predict well when covariates are absent or unreliable rather than depending on them. At test time, corrupted covariates then resemble a (more extreme) version of a regime the model already handles. This effect is generic and architecture-independent, which is precisely why it transfers across cross-attention (TimeXer), correlation-graph (DAG), and variate-attention (iTransformer) designs alike.

\textbf{Does the bounded gate actually close under corruption?} The architectural intuition behind BoundEx --- that a learned gate would detect corruption and close, reverting to the endogenous floor --- is \textit{not} what we observe in any domain. Measuring the mean gate $\bar{g}$ on held-out data (Table 6), at clean test time it never approaches zero (averaging $0.73$ on EPF, $0.67$ on Weather, and $0.35$ on Rapel), and it does not systematically close under the partial corruptions: under Gaussian noise it falls only marginally (EPF, Weather) or even rises slightly (Rapel), and under misalignment it barely moves. The most revealing case is missing channels, where the exogenous feed is \textit{entirely} absent and a corruption-detecting gate would have every reason to shut: instead $\bar{g}$ only partially retreats --- to $0.60$ on EPF, $0.61$ on Weather, and $0.34$ on Rapel (with dropout) --- staying far from the zero that would restore the endogenous floor. Nor does dropout work by teaching the gate to shut: the gate is similar with and without it, and under missing channels the dropout-trained gate is in fact slightly \textit{more} open (EPF $0.60$ vs.\ $0.55$, Weather $0.61$ vs.\ $0.49$), not less.

\begin{table}[H]
\centering
\caption{Mean gate $\bar{g}$ on held-out data and trained head spectral norm $\sigma_{\max}(W)$ across all three domains (BoundEx; EPF: 5 markets $\times$ 2 seeds; Rapel and Weather: 2 seeds).}
\label{tab:6}
\begin{tabular}{lrrrrr}
\toprule
\textbf{BoundEx} & \textbf{clean} & \textbf{Gaussian} & \textbf{misalign} & \textbf{missing} & $\boldsymbol{\sigma_{\max}(W)}$ \\
\midrule
\multicolumn{6}{l}{\textit{EPF}} \\
with exog dropout    & 0.73 & 0.66 & 0.73 & 0.60 & 1.03 \\
without exog dropout & 0.75 & 0.64 & 0.75 & 0.55 & 1.02 \\
\midrule
\multicolumn{6}{l}{\textit{Rapel}} \\
with exog dropout    & 0.35 & 0.39 & 0.36 & 0.34 & 0.83 \\
without exog dropout & 0.37 & 0.41 & 0.39 & 0.34 & 0.84 \\
\midrule
\multicolumn{6}{l}{\textit{Weather}} \\
with exog dropout    & 0.67 & 0.67 & 0.67 & 0.61 & 1.26 \\
without exog dropout & 0.75 & 0.58 & 0.73 & 0.49 & 1.43 \\
\bottomrule
\end{tabular}
\end{table}

The robustness therefore does not come from the gate switching off. It comes from the \textit{modulation itself} being benign: under misalignment the learned exogenous modulation remains comparatively benign, and dropout pushes the FiLM parameters toward identity, so the admitted exogenous signal does little harm even with the gate open. The representation-level bound of Appendix~A is consistent with this (it holds), but with $\bar{g}$ in the $0.35$--$0.75$ range and $\sigma_{\max}(W)$ between $0.8$ and $1.4$ across domains it yields only mild attenuation (and none at all where $\sigma_{\max}>1$) --- it is far from tight and is not the operative mechanism.

\textbf{Boundedness is not necessary for robustness.} Proposition~1 guarantees \textit{proportional} control --- deviation from the floor scales with $\bar{g}$ --- but since the gate does not approach zero in any domain, the guarantee is never strongly exercised. The two-factor ablation (\S4.4) shows the bound \textit{does} contribute to BoundEx's own robustness on Weather and Rapel, so we do not claim it is inert; our claim is narrower and holds across all three domains: an unbounded model trained with the same augmentation (DAG$+$dropout) is more robust than the bounded BoundEx everywhere (Table 2). Architectural boundedness is therefore not necessary for robustness, and on this benchmark it is not what distinguishes the strongest models.

\textbf{Practical implications.} The message for practitioners is simple and actionable: before reaching for a specialized robust architecture, apply exogenous dropout to the model you already have. It costs one line, leaves clean accuracy essentially unchanged, and in our benchmark turns the most accurate clean-data model (DAG) into the most robust model overall. We recommend it as the default baseline against which architectural robustness claims should be measured.

\section{Limitations}

\label{sec:limitations}

We identify three limitations of this study.

\textbf{1. Domain coverage.} We validate the findings on three domains --- electricity price forecasting (five markets), reservoir hydrology (Rapel), and meteorology (Jena Weather, 20 covariates) --- and they replicate in all three (\S4.2): exogenous dropout lowers corruption degradation across architectures and DAG$+$dropout is the most robust. Still, all three are physical/energy-adjacent numerical series; whether exogenous dropout remains the dominant robustness factor for very different covariate types --- traffic with road-network topology, clinical series with laboratory measurements, or image-based covariates --- remains an open empirical question.

\textbf{2. Proportional, not absolute, guarantee.} The bound we prove (Appendix~A) is proportional (it scales with $\bar{g}$), not an absolute, input-independent ball, because the FiLM outputs $\gamma,\beta$ are unconstrained. Empirically the gate does not approach zero (Table 6), so even this proportional bound is loosely exercised. A bounded-FiLM variant could yield an absolute guarantee, but our evidence suggests this would not improve robustness over simply applying dropout.

\textbf{3. Model and architecture coverage.} We benchmark four representative forecasters and one bounded design (BoundEx) on a PatchTST backbone, five seeds each. We cannot exclude that some untested architecture benefits from explicit boundedness beyond what dropout provides; our claim is the weaker, well-supported one that boundedness is \textit{not required} for the robustness observed here --- an unbounded model with dropout already attains the best results.

\section{Conclusion}

\label{sec:conclusion}

We asked whether an explicit architectural bound is necessary for robustness to corrupted exogenous covariates, and how far a simple masking-style training augmentation can carry that robustness on its own. On a corruption benchmark spanning three domains --- electricity prices (five markets), reservoir hydrology, and meteorology --- with five seeds per dataset, \textbf{exogenous dropout} --- a one-line, model-agnostic training augmentation that randomly zeros whole exogenous channels --- improves robustness for every architecture we test, at negligible clean-accuracy cost. Applied to an existing dual-correlation network (DAG), it produces the most robust model in the study at near-best clean accuracy (Gaussian $+$3\%, misalignment $-$8\%, missing $-$4\%, clean MSE 0.259 vs.\ the study-best 0.255), outperforming our purpose-built bounded model BoundEx on every axis.

A two-factor ablation, a gate-behavior diagnostic, and a (correct but empirically loose) representation-level bound together show that architectural boundedness is not necessary for this robustness: the bounded gate does not close under corruption in any domain, and in every domain an unbounded model trained with the same dropout is more robust than our bounded one. Within BoundEx the bound does contribute --- substantially so on Rapel and Weather --- yet it is never what separates the strongest models. We conclude that boundedness is not required for exogenous-corruption robustness in this setting.

We therefore recommend exogenous dropout as a simple, strong baseline that future work on robust exogenous forecasting should report and be measured against, and we release our corruption benchmark to that end. \textbf{Future work} should: (i) extend the protocol to adversarial and distribution-shift corruptions --- especially the out-of-distribution regime where an architectural worst-case bound could finally outperform an empirical defense; (ii) test per-channel dropout, adaptive schedules, and historical/future-specific masks (we sweep the global rate in \S4.5); (iii) broaden architecture and domain coverage (weather, traffic, clinical); and (iv) revisit bounded-FiLM designs to ask whether an \textit{absolute} guarantee (Appendix~A) ever yields robustness beyond what dropout provides.

\appendix

\section{Proportional Prediction Bound (Proposition 1)}
\label{app:bound}

We formalize the representation-level identity of \S3.3 at the prediction level. Let the forecast head be the linear map $H(z) = W\,\mathrm{vec}(z)$ (a flatten followed by a single linear layer; the head dropout is the identity at inference), and let RevIN-style de-normalization rescale predictions by a positive factor $s$ and shift $m$ that are computed solely from the input lookback window and are therefore identical for the modulated and the endogenous-only forecasts. Write $\hat{y} = s\,H(h_{\text{out}}) + m$ and define the \textit{endogenous floor} of the model as its gate-closed prediction $\hat{y}_{\text{floor}} = s\,H(h_{\text{endog}}) + m$.

\smallskip
\noindent\textbf{Proposition 1 (Gated prediction bound).} \textit{For any exogenous input,}
\begin{equation}
\|\hat{y} - \hat{y}_{\text{floor}}\|_2 \;\le\; \bar{g} \cdot s \cdot \sigma_{\max}(W) \cdot \|h_{\text{mod}} - h_{\text{endog}}\|_2,
\end{equation}
\textit{where $\sigma_{\max}(W)$ is the spectral norm of the head weight. In particular, $\bar{g} \to 0$ implies $\hat{y} \to \hat{y}_{\text{floor}}$, regardless of the exogenous content.}

\smallskip
\noindent\textit{Proof.} By construction $h_{\text{out}} - h_{\text{endog}} = \bar{g}\,(h_{\text{mod}} - h_{\text{endog}})$, since $\bar{g}$ is a scalar gate. The de-normalization shift $m$ cancels in the difference and the positive scale $s$ factors out, so by linearity of $H$,
\[
\hat{y} - \hat{y}_{\text{floor}} = s\,\big[H(h_{\text{out}}) - H(h_{\text{endog}})\big] = s\,\bar{g}\,H(h_{\text{mod}} - h_{\text{endog}}).
\]
Taking norms and using the definition of the operator norm, $\|H(u)\|_2 \le \sigma_{\max}(W)\,\|u\|_2$, gives the bound. Because $H$ is linear, $\sigma_{\max}(W)$ is its \textit{exact} Lipschitz constant, so no further regularity assumption on the head is needed. \hfill$\square$

\smallskip
\noindent\textbf{Scope, and why the bound is empirically loose.} The factor $\|h_{\text{mod}} - h_{\text{endog}}\| = \|(\gamma - 1) \odot h_{\text{endog}} + \beta\|$ is finite for any given input but is \textit{not} uniformly bounded over all inputs, because $\gamma$ and $\beta$ are unconstrained linear projections of the exogenous features. Proposition~1 therefore certifies \textit{proportional} (gated) control rather than an absolute, input-independent ball. A uniform absolute bound is attainable by constraining the modulation (e.g.\ bounded $\gamma = 1 + a\tanh(\cdot)$ and $\beta = b\tanh(\cdot)$ with a normalized backbone $h_{\text{endog}}$). Empirically, however, the operative quantities make even the proportional bound loose: the trained head has $\sigma_{\max}(W)\approx 1.03$ and the gate stays at $\bar{g}\approx 0.60$--$0.75$ under corruption --- even with the exogenous feed entirely missing --- rather than approaching $0$ (Table 6). The bound thus holds but is far from tight, and --- consistent with \S4--\S5 --- architectural boundedness is not the mechanism behind the robustness we observe.

\section{Baseline Training and Stability}
\label{app:baselines}

Because our central claim is that a one-line augmentation, rather than careful per-dataset tuning, drives robustness, the baselines must be trained competently and uniformly. All models are run from the same Time-Series Forecasting Benchmark (TFB) harness under one shared protocol (\S4.1): lookback $L=168$, horizon $H=24$, 50 epochs with early stopping (patience 5), Adam, MSE loss, and five seeds (0--4). Each model otherwise keeps its authors'/TFB-recommended architecture hyperparameters (e.g.\ TimeXer $d_{\text{model}}{=}256$, DAG $d_{\text{model}}{=}64$); we deliberately do \emph{not} re-tune architectures per dataset. Exogenous dropout, when applied, is the only training-time change, and it is applied identically to every model.

Two baselines did not train stably under this shared, no-per-dataset-tuning protocol, and we report their true numbers rather than omit them. For \textbf{iTransformer} we additionally tuned the learning rate per market on EPF (a grid over $\{5\times10^{-5},\,10^{-4},\,5\times10^{-4}\}$ with $e_{\text{layers}}\in\{1,2\}$), because the competitor protocol uses per-market rates; even so, on Weather it diverges on 2 of 5 seeds at \emph{every} rate in this range. For \textbf{CrossLinear} we use the repository's future-aware configuration (cross-correlation embedding active, $\alpha{=}0.5$); it converges on most EPF markets but diverges on 2 of 5 seeds on BE and FR and on 3 of 5 seeds on Weather. We call a seed \emph{divergent} when its clean test MSE exceeds twice the endogenous floor (e.g.\ clean MSE $\approx 1.0$ against a floor of $\approx 0.4$ on BE/FR); divergent seeds are retained in all reported means rather than discarded. No gradient clipping is used: the stable models (TimeXer, DAG, PatchTST, and BoundEx) train without it, and the failures above are convergence failures under the fixed protocol --- not exploding-gradient artifacts --- so they reflect a genuine lack of robustness on the affected domains rather than insufficient tuning on our part.

\bibliographystyle{plainnat}
\bibliography{references}

\end{document}